\pdfoutput=1

\documentclass[11pt]{article}

\usepackage{emnlp2021}

\usepackage{times}
\usepackage{latexsym}

\usepackage[T1]{fontenc}

\usepackage[utf8]{inputenc}

\usepackage{microtype}

\usepackage{graphicx}
\usepackage{amsthm}
\usepackage{amsmath}
\usepackage{subcaption}
\usepackage{booktabs}
\usepackage{amssymb}
\usepackage{multirow}
\usepackage{bm}
\usepackage{color}
\usepackage{verbatim}
\usepackage{enumerate}

\newcommand{\tabincell}[2]{\begin{tabular}{@{}#1@{}}#2\end{tabular}}
\definecolor{DarkRed}{RGB}{139,0,0}

\title{Progressive Adversarial Learning for Bootstrapping: A Case Study on Entity Set Expansion}

\author{Lingyong Yan${}^{1,3}$, Xianpei Han${}^{1,2}$, Le Sun${}^{1,2,*}$\\
${}^{1}$Chinese Information Processing Laboratory ~ ${}^{2}$State Key Laboratory of Computer Science\\
  Institute of Software, Chinese Academy of Sciences, Beijing, China \\
  ${}^{3}$University of Chinese Academy of Sciences, Beijing, China \\
  {\tt \{lingyong2014, xianpei, sunle\}@iscas.ac.cn}}

\begin{document}
\maketitle

\begin{abstract}

Bootstrapping has become the mainstream method for entity set expansion. Conventional bootstrapping methods mostly define the expansion boundary using seed-based distance metrics, which heavily depend on the quality of selected seeds and are hard to be adjusted due to the extremely sparse supervision. In this paper, we propose BootstrapGAN, a new learning method for bootstrapping which jointly models the bootstrapping process and the boundary learning process in a GAN framework. Specifically, the expansion boundaries of different bootstrapping iterations are learned via different discriminator networks; the bootstrapping network is the generator to generate new positive entities, and the discriminator networks identify the expansion boundaries by trying to distinguish the generated entities from known positive entities. By iteratively performing the above adversarial learning, the generator and the discriminators can reinforce each other and be progressively refined along the whole bootstrapping process. Experiments show that BootstrapGAN achieves the new state-of-the-art entity set expansion performance.
\end{abstract}
{
  \renewcommand{\thefootnote}{\fnsymbol{footnote}}
  \footnotetext[1]{Corresponding author.}
}
  
  \section{Introduction}
  \noindent
  Bootstrapping is a fundamental technique for entity set expansion (ESE).
  It starts from a few seed entities (e.g., \{\texttt{London, Beijing, Paris}\}) and iteratively extracts new entities in the target category (e.g., \{\texttt{Berlin, Moscow, Tokyo}\}) to expand the entity set, where new entities are often evaluated by their context similarities to seeds (e.g.,  sharing the same context pattern--``* is an important city'')~\cite{riloff_learning_1999,gupta_improved_2014,yan_end_2020}.
  During the above process, it is core to decide whether the new entities belong to the target category (within the expansion boundary) or not (outside the expansion boundary)~\cite{shi_probabilistic_2014,gupta_improved_2014}.
  
  \begin{figure}[!tbp]
    \centering
    \includegraphics[width=\columnwidth]{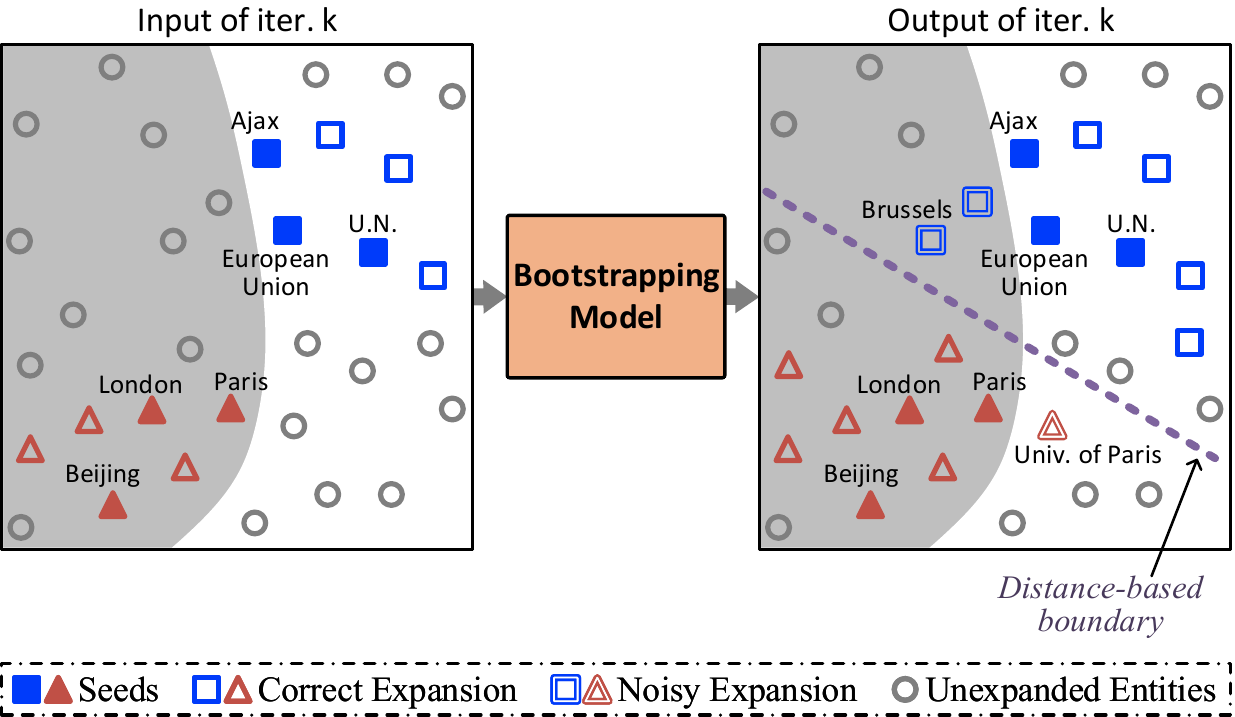}
    \caption{The expansion boundary problem of the bootstrapping technique. The areas in different background colors belong to different categories. Using the distance to positive entities will easily result in a bad expansion boundary at each iteration. }
    \label{fig:0}
\end{figure}
  
  However, it is challenging to determine the expansion boundaries during the whole bootstrapping process, since only several seeds are used as the supervision at the beginning.
  Firstly, it is obviously not enough to define a good boundary using only several positive entities.
  For example, as shown in Figure~\ref{fig:0}, when only using several positive entities to learn distance-based boundaries, the boundaries are usually far from optimum, which in turn influences the quality of following bootstrapping iterations.
  Therefore, it is critical to enhancing the boundary learning with more supervision signals or prior knowledge~\cite{thelen_bootstrapping_2002,curran_minimising_2007}.
  Secondly, bootstrapping is a dynamic process containing multiple iterations.
Therefore, the boundary needs to be synchronously adjusted with the bootstrapping model, i.e., a good boundary should precisely restrict the current bootstrapping model from expanding negative entities.
  
  Currently, most bootstrapping methods define expansion boundary using seed-based distance metrics, i.e., determining whether an entity should be expanded by comparing it with seeds.
  For instance, \citet{riloff_learning_1999,gupta_improved_2014,batista_semi-supervised_2015} define the boundary using pattern matching statistics or distributional similarities. 
  Unfortunately, these heuristic metrics heavily depend on the selected seeds, making the boundary biased and unreliable~\cite{curran_minimising_2007,mcintosh_reducing_2009}.
  Although some studies extend them with extra constraints~\cite{carlson_coupled_2010} or manual participants~\cite{berger_visual_2018},
  the requirement of expert knowledge makes them ad-hoc and inflexible.
  Some studies try to learn the distance metrics~\cite{zupon_lightly-supervised_2019,yan_end_2020}, but they still suffer from weak supervision.
  Furthermore, because the bootstrapping model and the boundary are mostly learned separately, it is hard for these methods to synchronously adjust the boundary when the bootstrapping model updates.
  
  To address the boundary learning problem, we propose a new learning method for bootstrapping--BootstrapGAN, which defines expansion boundaries via learnable discriminator networks, and jointly models the bootstrapping process and the boundary learning process in the generative adversarial networks (GANs) framework~\cite{goodfellow_generative_2014}:
  
  (1) Instead of using unified seed-based distance metrics, we define the expansion boundaries of different bootstrapping iterations using different learnable discriminator networks, where each of them directly determines whether an entity belongs to the same category of seeds at each iteration.
  By defining boundaries using discriminator networks, our method is flexible to use different classifiers and learnable using different algorithms.
  
  (2) At each bootstrapping iteration, by modeling the bootstrapping network as the generator and adversarially learning it with a discriminator network, our method can effectively resolve the sparse supervision problem for boundary learning.
  Specifically, at each bootstrapping iteration, the generator is trained to select the most confusing entities;
  the discriminator learns to determine the selected entities as negative instances, and previously expanded entities and seeds as positive instances.
  In this way, the generator and the discriminator can reinforce each other:
  the generator can enhance supervision signals for discriminator learning by selecting latent noisy entities, and the discriminator can influence the generator to select more indistinguishable entities.
  When reaching the generator-discriminator equilibrium, the discriminator finally learns a good expansion boundary that accurately identifies new entities, and the bootstrapping network can expand new positive entities within the boundaries.
  
  (3) By iteratively performing the above adversarial learning process, the bootstrapping network and the expansion boundaries are progressively refined along bootstrapping iterations.
  Specifically, we use a discriminator sequence containing multiple discriminators to progressively learn expansion boundaries for different bootstrapping iterations.
  And the bootstrapping network is also refined and restricted along the whole bootstrapping process by the current discriminator and previously learned discriminators.
  
  We conduct experiments over two datasets, and our BootstrapGAN achieves the new state-of-the-art performance for entity set expansion.
  
  
  \section{Progressive Adversarial Learning for Bootstrapping}
  
  
  In this section, we introduce our boundary learning method for bootstrapping models--BootstrapGAN (see Figure~\ref{fig:1}), which contains a generator--the bootstrapping network that performs the bootstrapping process, and a set of discriminators that determine the expansion boundaries for different bootstrapping iterations.
  The bootstrapping network and the discriminator networks are progressively and adversarially trained during the bootstrapping process.

  \subsection{Generator: Bootstrapping Network}
  \label{sec:generator}
  The generator is the bootstrapping model, which iteratively selects new entities to expand seed sets.

  We adopt the recently proposed end-to-end bootstrapping network--BootstrapNet~\cite{yan_end_2020} as the generator, which follows the encoder-decoder architecture:

  \begin{figure*}[!thbp]
    \centering
    \includegraphics[width=0.7\textwidth]{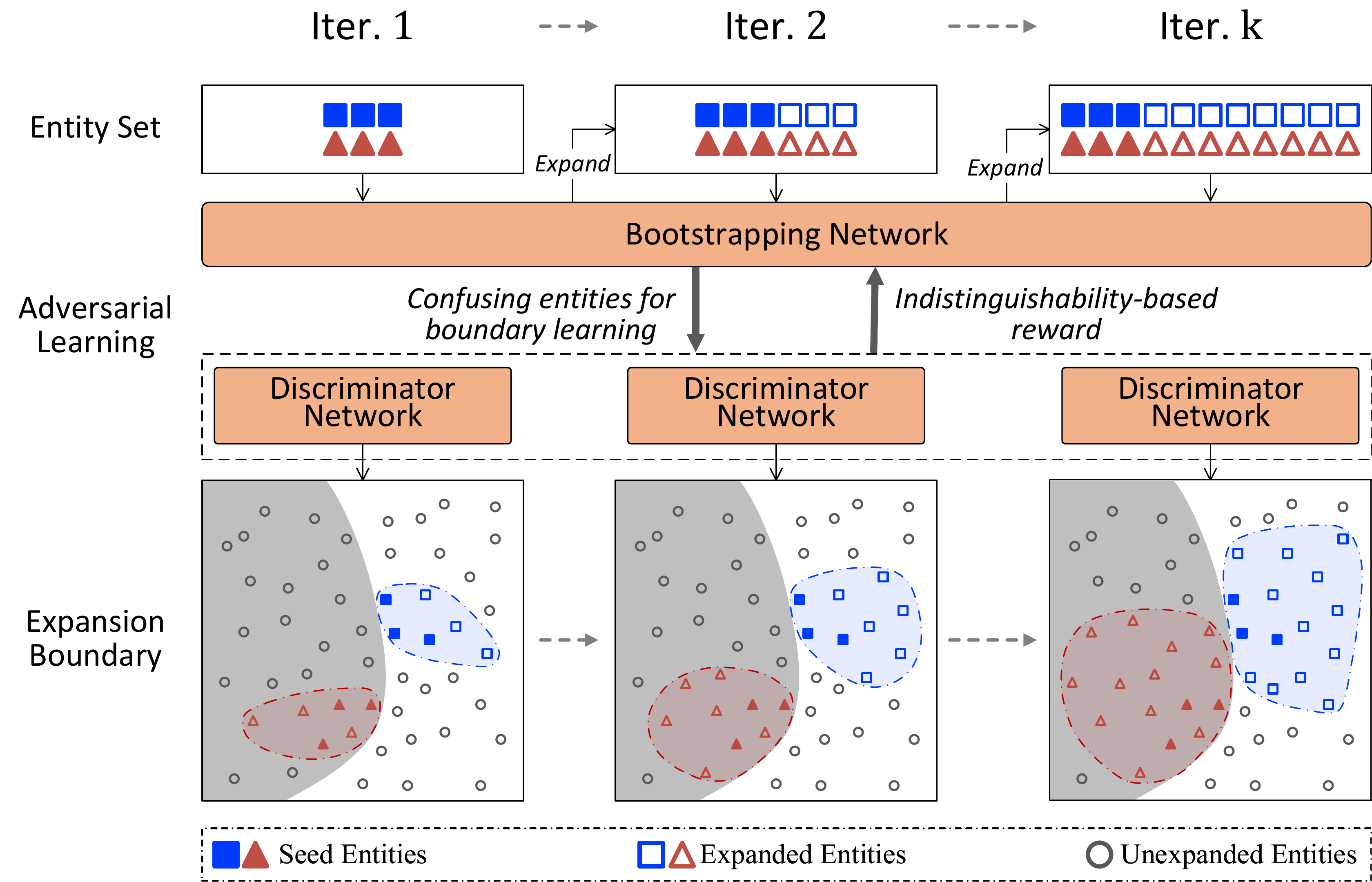}
    \caption{The overall framework of BootstrapGAN. }
    \label{fig:1}
\end{figure*}

  \paragraph{Encoder}
  The encoder is a multi-layer graph neural network (GNN) that encodes the context features around entities/patterns into their embeddings.
  And the encoder takes an entity-pattern bipartite graph as input to efficiently capture global evidence (i.e., the direct and multi-hop co-occurrences between entities and patterns).
  The bipartite graph is constructed from original datasets: entities and patterns are graph nodes; an entity and a pattern are linked if they co-occur. 

  Based on the above bipartite graph, each GNN layer aggregates information from node neighbors as follows:
  \begin{equation}
      v_i^l = \sigma(f(W^l v_i^{l-1}, \sum_{j\in N(i)}a_{i,j}^{l} W^l v_j^{l-1}))
      \label{eq:bootstrapnet_encoder}
  \end{equation}
  where $v_i^l$ is node $i$'s embedding after layer $l$, $N(i)$ are $i$'s neighbors, $W^l$ is the parameter matrix, $a_{i,j}^{l}$ is the attention-based weight, $f$ is a linear sum function, and $\sigma$ is the non-linear activation function.
  
  \paragraph{Decoder}
  After encoding entities and patterns, the GRU-based decoder sequentially generates new entities as the expansions, where each GRU step refers to one bootstrapping iteration.
  Specifically, the hidden state of the decoder represents the semantics of the target category.
  At each GRU step, the last expanded entities are used as the inputs to update the hidden state, which models the process that newly expanded entities are added to the current set, and therefore the set semantics should be updated (The first step inputs are seeds); then, the generating probabilities of a new entity are calculated as follows\footnote{This probability function is different from the original version of BootstrapNet that leverages cosine similarities.}:
  \begin{equation}
    \begin{split}
        P_k(j) &= \frac{exp(v_j M^{\mathsf{T}} h^k)}{\sum_{j'} exp(v_{j'} M^{\mathsf{T}} h^k)} \\
    \end{split}
    \label{eq:probability}
\end{equation}
\noindent where $h^k$ is the hidden state at $k$-th GRU step, $v_j$ is entity $j$'s embedding outputted by the encoder, $j'$ is a candidate entity, $M$ is the parameter matrix. And top-$N$ new entities are expanded at each step.
  
\subsection{Discriminator: Expansion Boundary}
Given positive entities (i.e., seeds and expanded entities), 
the discriminator defines the expansion boundary of each bootstrapping iteration by identifying whether a new entity is positive (i.e., belonging to the same category as positive entities) or negative (otherwise).

Instead of using seed-based distance metrics~\cite{riloff_learning_1999,gupta_improved_2014}, 
we take different categories of seeds into consideration, and design the discriminators to directly predict which category a new entity belongs to.
The motivation comes from two aspects:
(1) By enforcing the discriminator directly discriminating whether a new entity is positive to any category of seeds, the discriminator can essentially possess the category boundary and is flexible to leverage more supervision signals except for seeds;
(2) According to the mutual exclusive assumption~\cite{curran_minimising_2007} (i.e., \emph{most entities usually belong to only one category}),
it is better to leverage different categories of seeds to alleviate noises and simultaneously learn their expansion boundaries.
  
Specifically, we set our discriminator a multi-class classifier, which contains a GNN followed by an MLP layer:
The GNN module takes the entity-pattern bipartite graph as input, and encodes context features into entity embeddings as Eq. \ref{eq:bootstrapnet_encoder};
The MLP layer followed by a softmax function outputs the entity's category probabilities, where each category refers to one kind of seed set. 
And a new entity is only regarded as positive to the category with the highest probability.
Besides, we set the GNN module as 1-layer to avoid model overfitting.
  
  \subsection{Progressive Adversarial Learning}
  \label{sec:AdvLearn}
  
  \noindent
  To learn the above generator and discriminator, we design the following progressive adversarial learning process:
  Before bootstrapping, we pre-train the generator for better convergence (\textbf{Pre-training});
  At each bootstrapping iteration, the discriminator is used to learn the expansion boundaries of this iteration, and is adversarially trained with the generator to reinforce each other (\textbf{Local adversarial learning}).
  Along the whole bootstrapping process, we progressively refine the generator with multiple discriminators by iteratively performing the above local adversarial learning (\textbf{Global progressive refining}).
  
  \subsubsection{Pre-training}
  
  \noindent
  Many previous studies have suggested that pre-training is important for learning convergence in GANs~\cite{li_learning_2018,qin_dsgan_2018}.
  This paper pre-trains the generator (i.e., the bootstrapping network), and uses the following two kinds of pre-training algorithms:
  (1) The multi-view learning algorithm~\cite{yan_end_2020}, where the generator is co-trained with an auxiliary network.
  (2) Self-supervised and supervised pre-training using external resources~\cite{yan_global_2020}.
  Note that, since the external resources are not always accessible, we use the first algorithm as our default setting and set the second one as an alternative.
  
  \subsubsection{Local Adversarial Learning}
  
  \noindent
  At each bootstrapping iteration, the discriminator and the generator are learned using the following adversarial goals:
  the generator tries to generate new positive entities;
  the discriminator should distinguish new entities from current positive entities.
  
  However, it is difficult to adopt standard GAN settings for our method:
  (1) The discriminator is a multi-class classifier rather than a binary classifier.
  (2) The generator outputs discrete entities rather than continuous values.
  To address the above issues, we use a Shannon entropy-based objective that is consistent with the discriminator, and the policy gradient algorithm to optimize the generator.
  
  \paragraph{Shannon entropy-based learning objective}
  To make our GAN settings consistent with the multi-class discriminator, we modify the adversarial goals inspired by~\citet{springenberg_unsupervised_2016}:
  The generator tries to generate new entities that are certainly predicted as the same category as known positive entities by the discriminator; The discriminator tries to be not fooled by certainly assigning categories to the known positive entities and keeping uncertain about the class assignment for newly generated entities.
  
  Based on the new goals, we design a Shannon entropy-based learning objective, where the category assignment uncertainty is represented by the Shannon entropy.
  Formally, at bootstrapping iteration $k$, we use the following adversarial objective to learn the generator $G$ and the discriminator $D$:
  \begin{equation}
      \begin{split}
          \min \limits_{G} \max \limits_{D} & -\mathbb{E}_{e\sim S^c \cup G^c_{<k}}\left[H(p_D(c|e))\right] \\
          & + \lambda \mathbb{E}_{e\sim S^c \cup G^c_{<k}}\left[ CE(c, p_D(c|e)) \right] \\
          & + \mathbb{E}_{e' \sim G^c_{k}} \left[H(p_D(c|e'))\right] \\
      \end{split}
      \label{eq:objective}
  \end{equation}
  \noindent where $c$ is a target category, $S^c$ is the corresponding seed set, $G^c_{<k}$ is the set of expanded entities before iteration $k$, entities in $S^c \cup G^c_{<k}$ are regarded as positive entities, $G^c_{k}$ is the set of newly expanded entities at step $k$, $H(p_D(c|e))$ is the discriminator prediction entropy for $e$, $CE(\cdot)$ is the cross-entropy term to assign right classes for positive entities, and $\lambda$ is a hyper-parameter (this paper sets $\lambda = 1$).
  The first two terms of Eq.~\ref{eq:objective} aim to maximize the class assignment probabilities (i.e., minimizing the uncertainty) of positive entities, and the third term aims to maximize the entropies (i.e., maximizing the uncertainty) of newly generated entities.

  And we sample the same size of newly generated entities as the positive entities to balance the above adversarial training process (We still select top-$N$ entities for inference as Section \ref{sec:generator}).
  
  
  \paragraph{Policy gradient learning for generator}
  To optimize the generator that outputs discrete entities, we adopt the policy gradient algorithm.
  Specifically, we first rewrite the objective of the generator (the third term in Eq.~\ref{eq:objective}) as maximizing the following function (denoted as $L_G$):
  \begin{equation}
      \begin{split}
        L_G & = \mathbb{E}_{e \sim G^c_{k}} \left[-H(p_D(c|e)) \right] \\
                      & = \sum_{e \sim G^c_{k} } p_{G^c_{k}}(e)[-H(p_D(c|e))]
      \end{split}
      \label{eq:policy}
  \end{equation}
  where $p_{G^c_k}(e)$ is the expansion probability for entity $e$ at step $k$, and $e$ is a sampled discrete entity.
  We adopt the REINFORCE algorithm~\cite{williams_simple_1992} to directly calculate $L_G$'s gradient $\nabla_{\theta} L_G$ as:
  \begin{equation}
    \small
    \begin{split}
      \nabla_{\theta} L_G  &= \sum_{e' \sim G^c_{k}} p_{G^c_{k}}(e) \nabla_{\theta} \log p_{G^c_{k}}(e) R(e) \\
      R(e) &= p_{D}(c|e) - b
    \end{split}
      \label{eq:policy_gradient}
  \end{equation}
  where $p_{D}(c|e)$ is the probability of $e$ belonging to category $c$ returned by the discriminator, $R(e)$ is the indistinguishability-base reward for generator learning\footnote{Maximizing the probability of one class still equals maximizing the minus entropy (i.e., indistinguishability), thus we use the probability for efficiency.}, $b$ is the baseline value (This paper sets $b=\frac{1}{|C|}$, $|C|$ is the category number).
  
  \subsubsection{Global Progressive Refining}
  \noindent
  The local adversarial learning optimizes the generator and the discriminator at each bootstrapping iteration.
  This section describes how to refine them along the whole bootstrapping process--we call it global progressive refining.
  
  One naive refining method is to iteratively perform the above local adversarial learning using one generator and one discriminator.
  However, this setting is not suitable for the dynamic bootstrapping process.
  Firstly, since the positive entities are iteratively expanded, the expansion boundaries at sibling iterations should also be slightly different.
  Therefore, it is necessary to use different discriminators for different iterations.
  Secondly, for the end-to-end bootstrapping network~\cite{yan_end_2020}, restricting the outputs of the current iteration will influence the outputs of previous iterations,
  but the naive refining method cannot continuously restrict the expansions of previous iterations to already learned boundaries.
  
  Therefore, we propose a global progressive refining mechanism using a discriminator sequence containing multiple discriminators rather than one discriminator.
  Specifically:
  
  (1). For each bootstrapping iteration, we use a unique discriminator to learn its expansion boundaries. 
  That means for a total of $K$ bootstrapping iterations, the discriminator sequence contains $K$ different discriminators.
  
  (2). At the $k$-th iteration, discriminator $D_k$ is initialized by learned discriminator $D_{k-1}$; then $D_k$ and the generator $G$ are trained using the local adversarial learning until coverage; finally, $D_k$ can accurately define the expansion boundaries of iteration $k$ and keeps fixed in the following iterations.
      Through the above process, we can progressively refine the expansion boundaries by iteratively fitting new discriminators from previously learned boundaries to new ones.
  
  (3). At the $k$-th iteration, to restrict the generator's previous expansion to the learned boundaries (possessed by $\{D_1, D_2,...,D_{k-1}\}$), we also use the learned discriminator $D_i$ ($i \leq k $) to assign prediction probabilities as rewards for expanded entities at iteration $i$. Finally, we replace the generator's gradient calculated by Eq.~\ref{eq:policy_gradient} as:
  {
    \setlength\abovedisplayskip{1pt}
      \begin{equation}
        \small
        \nabla_{\theta} L_G = \sum_{i=1}^k \sum_{e' \sim G^c_{i}} p_{G^c_{i}}(e) \nabla_{\theta} \log p_{G^c_{i}}(e) [p_{D_i}(c|e) - b]
      \end{equation}
  }
\noindent where $D_k$ is the discriminator to be learned at iteration $k$, and $\{D_1,...,D_{k-1}\}$ are already learned discriminators.
  
  \section{Experiments}

  \begin{figure*}[!tp]
    \setlength{\abovecaptionskip}{0.5em}
    \setlength{\belowcaptionskip}{0em}
      \begin{subfigure}{0.5\textwidth}
          \includegraphics[width=\textwidth]{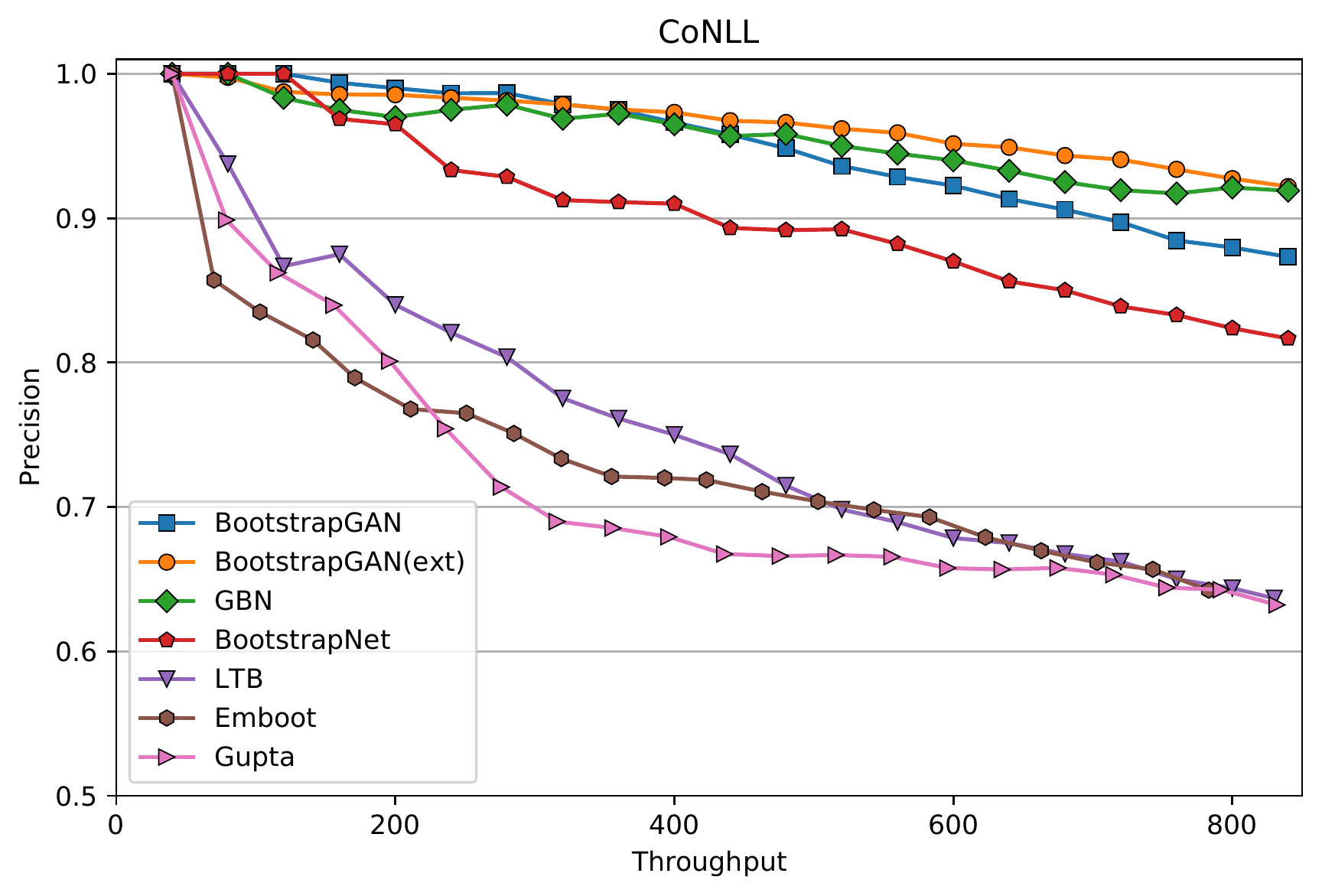}
      \end{subfigure}%
      \begin{subfigure}{.5\textwidth}
          \includegraphics[width=\textwidth]{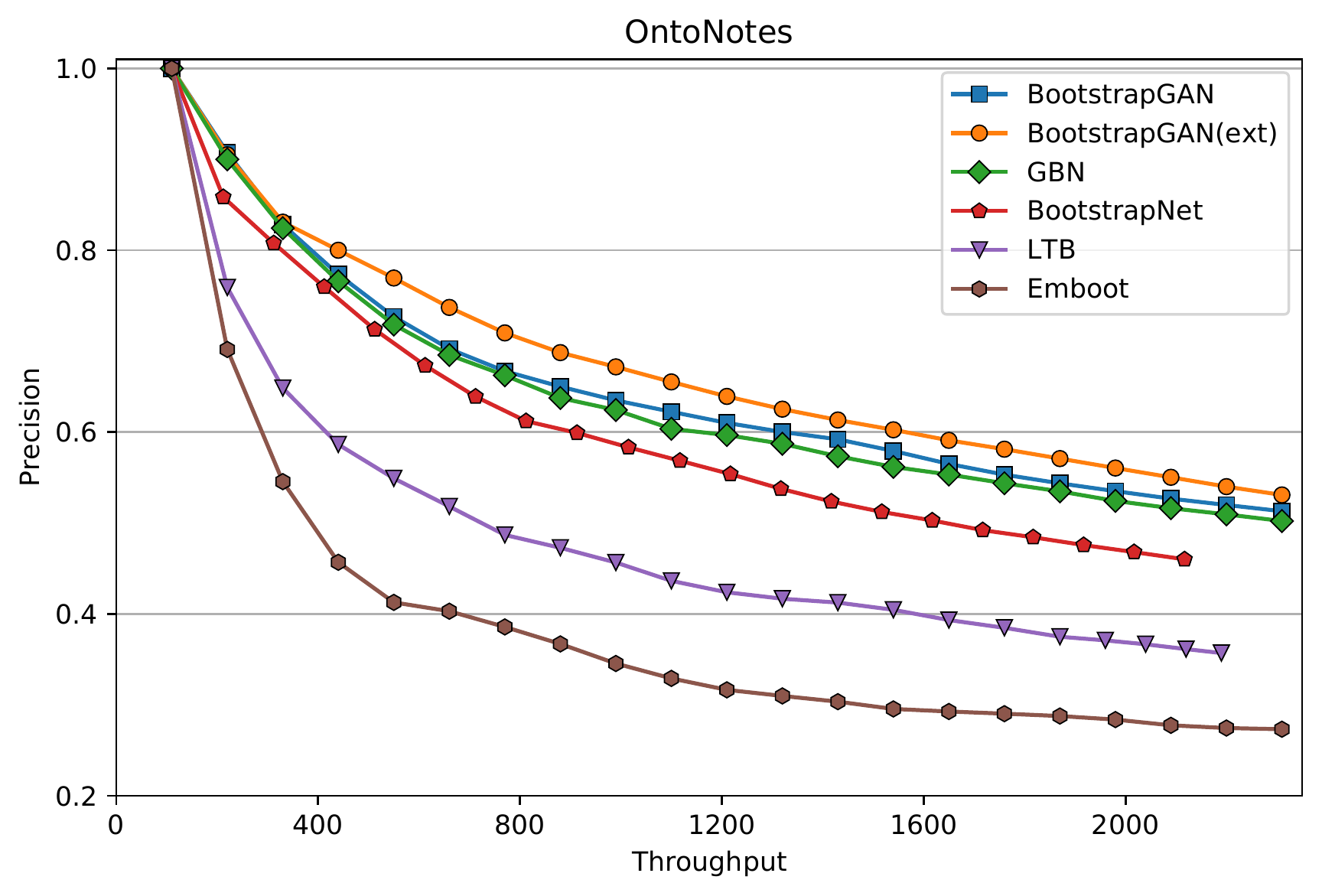}
      \end{subfigure}
      \caption{The precision-throughput curves on CoNLL and OntoNotes.}
      \label{fig:main}
  \end{figure*}
  
  \subsection{Experimental Setup}
  
  \begin{table}[thbp]
      \setlength{\belowcaptionskip}{-1.em}
      \centering
      \resizebox*{0.7\linewidth}{!}{
          \begin{tabular}{cc}
          \toprule
          \textbf{Hyper-parameter} & \textbf{Value} \\
          \midrule
          Learning Rate & 1e-4 \\
          Weight Decay & 1e-3 \\
          Dropout Rate & 0.1 \\
          Training Epoch per Iteration & 10 \\
          \bottomrule
          \end{tabular}%
      }
          \caption{Main hyper-parameter settings.}
      \label{tab:parameter}%
  \end{table}%

\begin{table*}[!tbp]
  \setlength{\belowcaptionskip}{-1em}
    \centering
    \resizebox*{0.95\linewidth}{!}{
      \begin{tabular}{ccccccccc}
      \toprule
      \multicolumn{1}{c}{\multirow{2}[4]{*}{\textbf{Method}}} & \multicolumn{4}{c}{\textbf{CoNLL}} & \multicolumn{4}{c}{\textbf{OntoNotes}} \\
      \cmidrule(lr){2-5} \cmidrule(lr){6-9}          & \multicolumn{1}{c}{\textbf{P@5}} & \multicolumn{1}{c}{\textbf{P@10}} & \multicolumn{1}{c}{\textbf{P@20}} & \multicolumn{1}{c}{\textbf{Mean}} & \multicolumn{1}{c}{\textbf{P@5}} & \multicolumn{1}{c}{\textbf{P@10}} & \multicolumn{1}{c}{\textbf{P@20}} & \multicolumn{1}{c}{\textbf{Mean}} \\
      \midrule
      Gupta & 70.4  & 63.4  & 61.3  & 65.0  & - & - & - & - \\
      LTB   & 78.5  & 71.0  & 62.2  & 70.6  & 42.2  & 36.6  & 32.3  & 37.0  \\
      \midrule
      Emboot & 71.3  & 68.8  & 62.3  & 67.5  & 28.4  & 24.8  & 23.7  & 25.6  \\
      BootstrapNet & 92.0  & 88.3  & 80.8  & 87.0  & 60.2  & 52.1  & 43.1  & 51.8  \\
      GBN   & 97.0  & 95.3  & 91.5  & 94.6  & 62.2  & 55.6  & 47.7  & 55.2  \\
      \midrule
      BootstrapGAN & \textbf{98.7($\pm$0.5)} & 94.8($\pm$0.4)  & 86.4($\pm$0.9)  & 93.3  & 63.0($\pm$0.7)  & 57.1($\pm$0.4)  & 48.9($\pm$0.5)  & 56.3  \\
      \ \ - pre-training & 98.0($\pm$0.4)  & 94.5($\pm$0.6)  & 87.1($\pm$0.5)  & 93.2  & 54.8($\pm$1.8)  & 49.1($\pm$1.3)  & 44.0($\pm$1.3)  & 49.3  \\
      \midrule
      BootstrapGAN(ext) & 98.0($\pm$0.5)  & \textbf{96.4($\pm$0.5)} & \textbf{91.8($\pm$0.9)} & \textbf{95.4} & \textbf{68.5($\pm$0.7)} & \textbf{60.3($\pm$0.5)} & \textbf{50.7($\pm$0.4)} & \textbf{59.8} \\
      \bottomrule
      \end{tabular}%
    }
    \caption{The P@K values (\%) of different bootstrapping models.}
    \label{tab:all_results}%
  \end{table*}%
  
  \paragraph{Datasets}
  The evaluation datasets we used are published by~\citet{zupon_lightly-supervised_2019} and used by~\citet{yan_end_2020}--CoNLL and OntoNotes:
  The CoNLL contains 4 categories (5,522 entities),
  and the OntoNotes contains 11 categories (19,984 entities).
  For each category, 10 entities are used as the seeds, and all {$n$}-grams ($n\leq4$) around candidate entities are defined as the context patterns.
  
  \paragraph{Baselines}
  We compare BootstrapGAN with the following baselines:
  
  (1) Bootstrapping methods using heuristic seed-based distance metrics, including statistical metric--\textbf{Gupta}~\cite{gupta_improved_2014}, and lookahead search-based method--\textbf{LTB}~\cite{yan_learning_2019};
  
  (2) Bootstrapping methods using weakly-supervised learned boundaries, including custom embedding-based method--\textbf{Emboot}~\cite{zupon_lightly-supervised_2019}, and end-to-end bootstrapping model learned by multi-view --\textbf{BootstrapNet}~\cite{yan_end_2020}, end-to-end bootstrapping model pre-trained using external datasets--\textbf{GBN}~\cite{yan_global_2020}.

  For BootstrapGAN, we report the results of its two versions: BootstrapGAN, which uses the multi-view learning algorithm for pre-training; BootstrapGAN(ext), which uses external datasets for pre-training like~\citet{yan_global_2020}.
  
  \paragraph{Evaluation Metrics}
  Following ~\citet{zupon_lightly-supervised_2019}, this paper uses the precision-throughput curves to compare all methods.
  For further precise evaluation, we also report the precision@$K$ values (P@K, i.e., the precision at expansion step $K$).
  And we run our method for 10 repetitive training pieces and report the mean values of P@K as well as the standard deviations.
  
  \paragraph{Implementation}
   We implement the BootstrapGAN using the PyTorch~\cite{paszke_pytorch_2019} with the PyTorch Geometric extension~\cite{fey_fast_2019}, and run it on a single Nvidia TiTan RTX GPU.
  And we use Adam~\cite{kingma_adam:_2015} and Rmsprop~\cite{tieleman_lecture_2012} to respectively optimize the generator and the discriminators.
  Main hyper-parameters are shown in Table \ref{tab:parameter}. 
  Our code is released at \url{https://www.github.com/lingyongyan/BootstrapGAN}.  
  
  \subsection{Overall Evaluation Results}
  
  \noindent
  The precision-throughput curves of all methods are shown in Figure~\ref{fig:main}, and P@K values are also shown in Table~\ref{tab:all_results}.
  We can observe that:

  (1) \textbf{Adversarial learning can effectively learn good expansion boundaries for bootstrapping models}.
  Comparing to all baselines without external resource pre-training (i.e., Gupta, LTB, Emboot, and BootstrapNet), BootstrapGAN achieves significant improvements (All p-values of t-test evaluation are less than 0.01), and the precision-throughput curves of BootstrapGAN are the most smooth ones.
  That means more correct entities and less noisy entities are expanded at each iteration. 
  It verifies that the learned expansion boundaries of BootstrapGAN contain fewer noisy entities than other methods, and therefore are the better boundaries.
  Besides, comparing to the baseline model using external resources for pre-training (i.e., GBN), the external resource pre-trained version--BootstrapGAN(ext) also outperforms it. 

  (2) \textbf{Progressive adversarial learning is complementary with self-supervised and supervised pre-training, and combining them can achieve the new state-of-the-art performance}.
  Comparing to the original BootstrapGAN, BootstrapGAN(ext), which combines self-supervised and supervised pre-training, achieves further improvements: On CoNLL, the P@10 and P@20 values achieve 1.6\% and 5.4\% improvements; On OntoNotes, the P@10 and P@20 values achieve 3.2\% and 1.8\% improvements.

  (3) \textbf{The end-to-end bootstrapping paradigm outperforms other bootstrapping methods}.
  Comparing to other methods, the end-to-end learning methods (i.e., BootstrapNet, GBN and BootstrapGAN, BootstrapGAN(ext)) can achieve obviously higher performance.
  And comparing to the BootstrapNet/GBN, BootstrapGAN/BootstrapGAN(ext) can further achieve noticeable improvements, especially on the more complex dataset–OntoNotes.

  
  \subsection{Detail Analysis}

  \paragraph{Effect of pre-training strategies.}
  To analyze the effects of pre-training, we compare the performance of BootstrapGAN using different pre-training settings (see Table~\ref{tab:all_results}):
  \textbf{BootstrapGAN}, and BootstrapGAN without pre-training (\textbf{- pre-train}).
  And we can see that: pre-training is an effective way to improve bootstrapping performance in some tasks.
  Without the pre-training, the BootstrapGAN's performance on OntoNotes substantially drops--all mean P@K values decrease at least 4.9\%.
  This may be because complex datasets (e.g., the OntoNotes) usually contain massive amounts of entities, and the search space of the bootstrapping network is extremely large, which makes it hard to converge to the optimum without appropriate pre-training.

  \begin{table}[!tbp]
    \setlength{\belowcaptionskip}{-1em}
    \centering
    \small
    \resizebox*{\linewidth}{!}{
    \begin{tabular}{ccccccc}
        \toprule
        \multirow{2}[3]{*}{{\tabincell{c}{\textbf{Refining}\\\textbf{Strategy}}}} & \multicolumn{3}{c}{\textbf{CoNLL}}    & \multicolumn{3}{c}{\textbf{OntoNotes}} \\
    \cmidrule(lr){2-4} \cmidrule(lr){5-7}          & \textbf{P@5} & \textbf{P@10} & \textbf{P@20} & \textbf{P@5} & \textbf{P@10} & \textbf{P@20} \\
        \midrule
        BootstrapGAN & \textbf{98.7} & \textbf{94.8} & 86.4 & \textbf{63.0} & \textbf{57.1} & \textbf{48.9} \\
        - refining & 93.1  & 83.0  & 73.1  & 56.4  & 50.7  & 43.2  \\
        - g-refining & 95.2  & 92.6  & \textbf{87.0} & 63.0  & 56.4  & 48.5 \\
        \bottomrule
    \end{tabular}%
    }
    \caption{Performance comparision of BootstrapGAN with different refining mechanisms.}
    \label{tab:strategies}%
\end{table}%
  
  \paragraph{Effect of global progressive refining.}
  To analyze the effects of global progressive refining, we conduct the comparison experiments with different refining mechanisms (see Table~\ref{tab:strategies}):
  original settings using global progressive refining (\textbf{BootstrapGAN});
  performing local adversarial learning without refining, i.e., only seeds are taken as positive entities, all expanded entities from different iterations are taken as negative ones in Eq.~\ref{eq:objective}(\textbf{- refining});
  performing refining using the naive refining mechanism rather than our global progressive refining (\textbf{- g-refining}).
  From Table \ref{tab:strategies}, we can see that:
  
  (1) Refining is useful when performing adversarial learning for bootstrapping.
  Without the refining mechanism, the BootstrapGAN performance sharply drops on both datasets (All P@K values decrease by at least 5.6\%).
  
  (2) Our global progressive refining mechanism is very suitable for BootstrapGAN learning.
  By replacing the global progressive refining with the naive mechanism, we can see that most BootstrapGAN performance results decrease, especially on P@5 and P@10.
  This verifies our observation that the expansion of previous iterations can be influenced when adversarially learning for later iterations.
  And our global progressive refining can well alleviate the influence, and therefore a better refining mechanism.

  \begin{figure}[!t]
    \setlength{\belowcaptionskip}{-1em}
    \centering  
    \includegraphics[width=\linewidth]{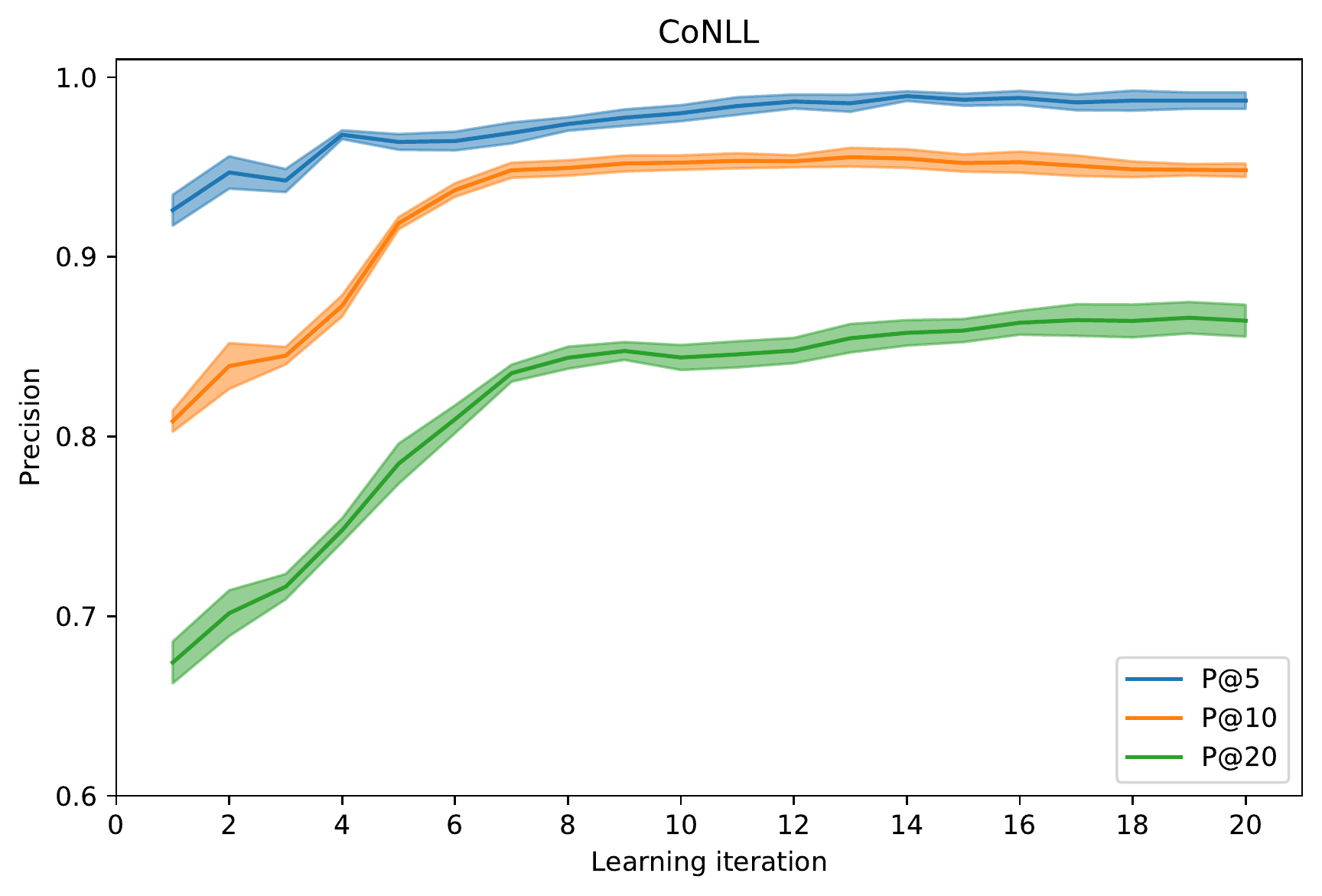} \\
    \includegraphics[width=\linewidth]{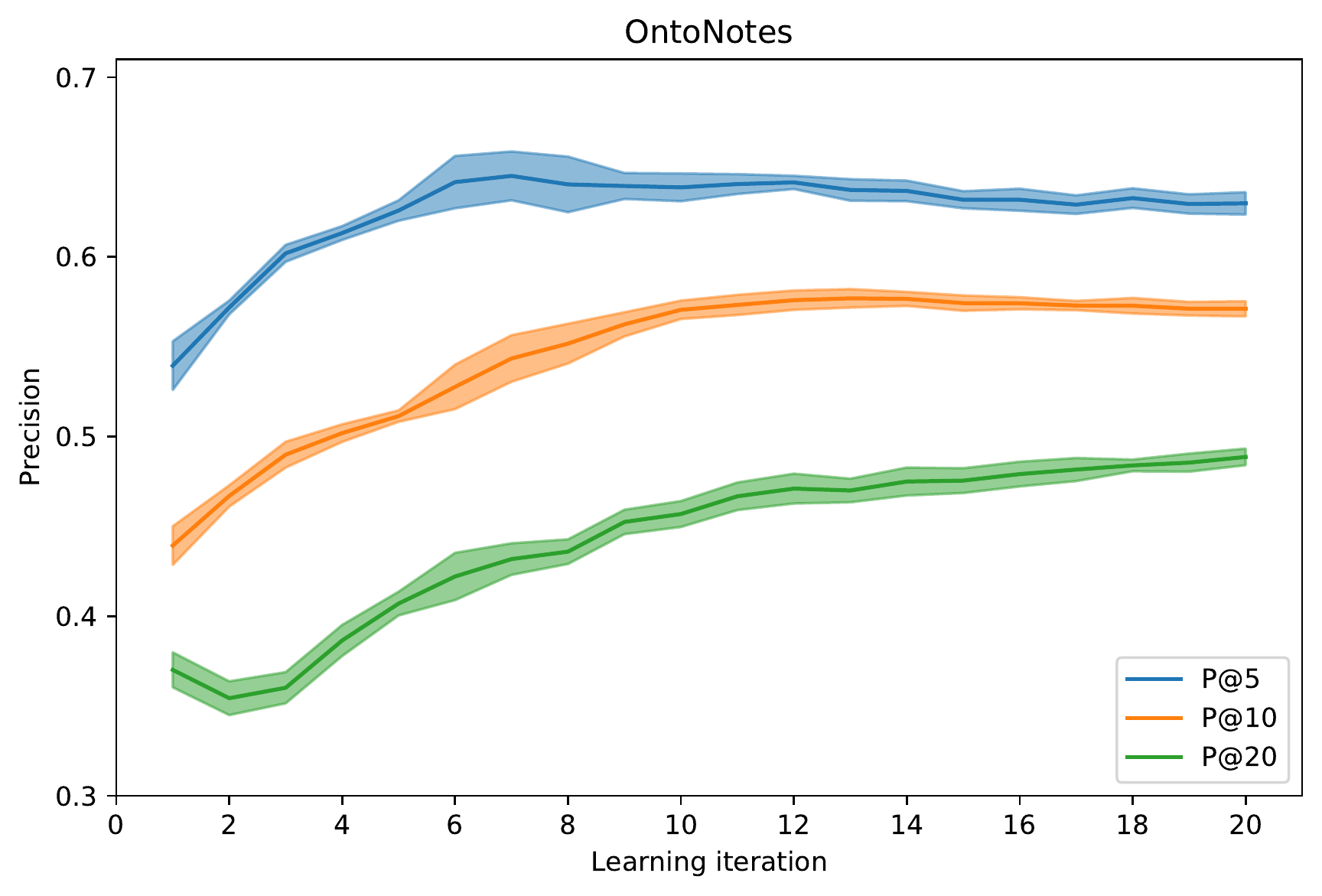}
    \caption{Mean and standard deviation bands of P@K values of BootstrapGAN across different learning iterations.}
    \label{fig:learning}
  \end{figure}

  \begin{table*}[!t]
    \setlength{\abovecaptionskip}{0.5em}
    \setlength{\belowcaptionskip}{-1em}
  \centering
  \resizebox*{\textwidth}{!}{
  \begin{tabular}{p{5em}p{25em}p{25em}}
      \toprule
      \multicolumn{1}{c}{Iter.} & \multicolumn{1}{c}{\textbf{BootstrapNet}} & \multicolumn{1}{c}{\textbf{BootstrapGAN}} \\
      \midrule
      \multicolumn{1}{c}{1} & the States,  \textcolor{DarkRed}{Asia Development Corp.}, \textcolor{DarkRed}{Atlantic}, \textcolor{DarkRed}{North China Area Army}, \textcolor{DarkRed}{Continent}, \textcolor{DarkRed}{the Gulf of Mexico}, \textcolor{DarkRed}{Mediterranean}, \textcolor{DarkRed}{Scandinavia},  \textcolor{DarkRed}{the East Coast}, \textcolor{DarkRed}{Bank of China} & Mexico, Poland, Romania, Denmark,  Moscow, The Netherlands, Vienna, Hungary, Greece, Bulgaria \\
      \midrule
      \multicolumn{1}{c}{10} & Indonesia, Somalia, \textcolor{DarkRed}{Northern German}, \textcolor{DarkRed}{the Convention on Trade in Endangered Species}, \textcolor{DarkRed}{...},  \textcolor{DarkRed}{the Western Hemisphere}, \textcolor{DarkRed}{the Asia Pacific region} & Melbourne, Havana,  Tajikstan, Lausanne,  Tehran, Abidjan,  Thailand, Bahrain, Aqaba, \textcolor{DarkRed}{the Shaanxi International Exhibition Center} \\
      \midrule
      \multicolumn{1}{c}{20} & the Republic of Iraq, \textcolor{DarkRed}{the United Nations World Human Rights Convention}, \textcolor{DarkRed}{Arab - Israelis}, \textcolor{DarkRed}{...}, \textcolor{DarkRed}{Budget Group},  \textcolor{DarkRed}{Sino - Kirghizian} & Adelaide, Rangoon, Cologne, Madrid, Phnom Penh,  Jinan,  Karachi, Palermo,  \textcolor{DarkRed}{Baghdad Airport}, \textcolor{DarkRed}{the North Pole} \\
      \bottomrule
      \end{tabular}%
  }
      \caption{The examples of expanded GPEs using BootstrapNet and BootstrapGAN (Entities in red are noisy entities).}
  \label{tab:case_study}%
\end{table*}%
  
  \paragraph{Stability of adversarial learning.}
  To analyze the stability of our adversarial learning method, we report the P@K values of BootstrapGAN at different iterations (see Figure \ref{fig:learning}).
  We can see that:
  (1) Our adversarial learning method can coverage quickly.
  At around the 10th bootstrapping iteration, the performance of BootstrapGAN reaches a reasonable level.
  (2) Our adversarial learning method is stable.
  On both datasets, most P@K values steadily increase with more training iterations, and the standard deviations of most P@K values progressively decrease.
  Those can verify the stability of our learning algorithm
  (Although some P@K values decrease a little from iteration 10 to iteration 20, we still consider our algorithm stable since the differences are slight enough to be omitted).
  
  \paragraph{Examples for learned expansion boundaries.}
  To intuitively show the quality of learned expansion boundaries by BootstrapGAN, we show a typical case of different expanded entities for GPE (geopolitical entities) on the OntoNotes using BootstrapNet and BootstrapGAN (see Table \ref{tab:case_study})\footnote{The seeds are \{\texttt{Washington, New York, the United States, Russia, Iran, Hong Kong, France, London, California, China}\}.}.
  And we can see that BootstrapGAN can expand more correct entities, and most of them are tightly related to the GPE semantics;
  while the expansion boundaries of BootstrapNet contain many noisy entities at the very beginning and tend to introduce more noises at later iterations.
  This further verifies the importance of expansion boundary learning and BootstrapGAN's effectiveness.

  \section{Related Work}
  \paragraph{Bootstrapping}
  Bootstrapping is a widely used technique for information extraction~\cite{riloff_automatically_1996,ravichandran_learning_2002,yoshida_person_2010,angeli_bootstrapped_2015,saha_bootstrapping_2017}, and also benefits many other NLP tasks, like question answering~\cite{ravichandran_learning_2002}, named entity translation~\cite{lee_bootstrapping_2013}, knowledge base population~\cite{angeli_bootstrapped_2015}, etc.
  To address the expansion boundary problem, most early methods~\cite{riloff_automatically_1996,riloff_learning_1999} heuristically decide boundaries using pattern-matching statistics, but often result in a rapid quality degrading, which is known as the semantic drifting~\cite{curran_minimising_2007}.
  To reduce semantic drifting, some studies leverage external resources or constraints, e.g., mutual exclusive constraints~\cite{yangarber_unsupervised_2002,thelen_bootstrapping_2002, curran_minimising_2007, carlson_coupled_2010}, lexical and statistical features~\cite{gupta_improved_2014}, lookahead feedbacks~\cite{yan_learning_2019}, manually defined patterns~\cite{zhang_empower_2020}.
  However, those heuristic constraints are usually not flexible due to their requirement for expert efforts.
  In contrast, recent studies focus on learning the distance metrics to determine boundaries using weak supervision~\cite{gupta_distributed_2015,berger_visual_2018,zupon_lightly-supervised_2019,yan_end_2020}.
  For example, ~\citet{yan_end_2020} propose an end-to-end bootstrapping network learned by multi-view learning, and extend it by self-supervised and supervised pre-training~\cite{yan_global_2020}.
  However, these methods usually learn a loose boundary using sparse supervision.
  Furthermore, these methods' boundary learning process and model learning process are usually separately performed and therefore fail to be adjusted synchronously.
  
  \textbf{Adversarial Learning in NLP}
  Adversarial learning ~\cite{goodfellow_generative_2014} is widely applied in NLP. For example,
  in sequential generation tasks, GAN is mainly used to alleviate the problem of lacking explicitly defined criteria~\cite{yu_seqgan_2017,lin_adversarial_2017,yang_improving_2018}.
  GAN has also been used in weakly supervised information extraction to identify informative instances and filter out noises~\cite{qin_dsgan_2018,wang_adversarial_2019}, which inspires our method.
  
\section{Conclusion}
Due to very sparse supervision and the dynamic nature, one fundamental challenge of bootstrapping is how to learn precise expansion boundaries.
In this paper, we propose an effective learning method for bootstrapping–BootstrapGAN, which defines expansion boundaries via learnable discriminator networks and jointly models the bootstrapping process and the boundary learning process in the GANs framework.
Experimental results show that, by adversarially learning and progressively refining the bootstrapping network and the discriminator networks, our method achieves the new state-of-the-art performance.
In the future, we plan to leverage extra knowledge (e.g., knowledge graph) to improve bootstrapping learning.

\section*{Acknowledgments}
This work is supported by the National Natural Science Foundation of China under Grants no. U1936207 and 61772505, Beijing Academy of Artiﬁcial Intelligence (BAAI2019QN0502), and in part by the Youth Innovation Promotion Association CAS(2018141).
  
{
\bibliographystyle{acl_natbib}
\bibliography{references}

\begin{thebibliography}{35}
\expandafter\ifx\csname natexlab\endcsname\relax\def\natexlab#1{#1}\fi

\bibitem[{Angeli et~al.(2015)Angeli, Zhong, Chen, Chaganty, Bolton, Premkumar,
  Pasupat, Gupta, and Manning}]{angeli_bootstrapped_2015}
Gabor Angeli, Victor Zhong, Danqi Chen, Arun~Tejasvi Chaganty, Jason Bolton,
  Melvin Jose~Johnson Premkumar, Panupong Pasupat, Sonal Gupta, and
  Christopher~D Manning. 2015.
\newblock Bootstrapped self training for knowledge base population.
\newblock In \emph{TAC}.

\bibitem[{Batista et~al.(2015)Batista, Martins, and
  Silva}]{batista_semi-supervised_2015}
David~S. Batista, Bruno Martins, and M{\'a}rio~J. Silva. 2015.
\newblock \href {https://doi.org/10.18653/v1/D15-1056} {Semi-supervised
  bootstrapping of relationship extractors with distributional semantics}.
\newblock In \emph{Proceedings of the 2015 Conference on Empirical Methods in
  Natural Language Processing}, pages 499--504, Lisbon, Portugal. Association
  for Computational Linguistics.

\bibitem[{Berger et~al.(2018)Berger, Nagesh, Levine, Surdeanu, and
  Zhang}]{berger_visual_2018}
Matthew Berger, Ajay Nagesh, Joshua Levine, Mihai Surdeanu, and Helen Zhang.
  2018.
\newblock \href {https://doi.org/10.18653/v1/D18-1229} {Visual supervision in
  bootstrapped information extraction}.
\newblock In \emph{Proceedings of the 2018 Conference on Empirical Methods in
  Natural Language Processing}, pages 2043--2053, Brussels, Belgium.
  Association for Computational Linguistics.

\bibitem[{Carlson et~al.(2010)Carlson, Betteridge, Wang, Jr., and
  Mitchell}]{carlson_coupled_2010}
Andrew Carlson, Justin Betteridge, Richard~C. Wang, Estevam R.~Hruschka Jr.,
  and Tom~M. Mitchell. 2010.
\newblock \href {https://doi.org/10.1145/1718487.1718501} {Coupled
  semi-supervised learning for information extraction}.
\newblock In \emph{Proceedings of the Third International Conference on Web
  Search and Web Data Mining}, pages 101--110. {ACM}.

\bibitem[{Curran et~al.(2007)Curran, Murphy, and
  Scholz}]{curran_minimising_2007}
James~R. Curran, Tara Murphy, and Bernhard Scholz. 2007.
\newblock Minimising semantic drift with mutual exclusion bootstrapping.
\newblock In \emph{PACLING}, pages 172--180.

\bibitem[{Fey and Lenssen(2019)}]{fey_fast_2019}
Matthias Fey and Jan~E. Lenssen. 2019.
\newblock Fast graph representation learning with {PyTorch Geometric}.
\newblock In \emph{ICLR Workshop on Representation Learning on Graphs and
  Manifolds}.

\bibitem[{Goodfellow et~al.(2014)Goodfellow, Pouget{-}Abadie, Mirza, Xu,
  Warde{-}Farley, Ozair, Courville, and Bengio}]{goodfellow_generative_2014}
Ian~J. Goodfellow, Jean Pouget{-}Abadie, Mehdi Mirza, Bing Xu, David
  Warde{-}Farley, Sherjil Ozair, Aaron~C. Courville, and Yoshua Bengio. 2014.
\newblock \href
  {https://proceedings.neurips.cc/paper/2014/hash/5ca3e9b122f61f8f06494c97b1afccf3-Abstract.html}
  {Generative adversarial nets}.
\newblock In \emph{Advances in Neural Information Processing Systems 27: Annual
  Conference on Neural Information Processing Systems 2014}, pages 2672--2680.

\bibitem[{Gupta and Manning(2014)}]{gupta_improved_2014}
Sonal Gupta and Christopher Manning. 2014.
\newblock \href {https://doi.org/10.3115/v1/W14-1611} {Improved pattern
  learning for bootstrapped entity extraction}.
\newblock In \emph{Proceedings of the Eighteenth Conference on Computational
  Natural Language Learning}, pages 98--108, Ann Arbor, Michigan. Association
  for Computational Linguistics.

\bibitem[{Gupta and Manning(2015)}]{gupta_distributed_2015}
Sonal Gupta and Christopher~D. Manning. 2015.
\newblock \href {https://doi.org/10.3115/v1/N15-1128} {Distributed
  representations of words to guide bootstrapped entity classifiers}.
\newblock In \emph{Proceedings of the 2015 Conference of the North {A}merican
  Chapter of the Association for Computational Linguistics: Human Language
  Technologies}, pages 1215--1220, Denver, Colorado. Association for
  Computational Linguistics.

\bibitem[{Kingma and Ba(2015)}]{kingma_adam:_2015}
Diederik~P. Kingma and Jimmy Ba. 2015.
\newblock \href {http://arxiv.org/abs/1412.6980} {Adam: {A} method for
  stochastic optimization}.
\newblock In \emph{3rd International Conference on Learning Representations}.

\bibitem[{Lee and Hwang(2013)}]{lee_bootstrapping_2013}
Taesung Lee and Seung-won Hwang. 2013.
\newblock \href {https://aclanthology.org/P13-1062} {Bootstrapping entity
  translation on weakly comparable corpora}.
\newblock In \emph{Proceedings of the 51st Annual Meeting of the Association
  for Computational Linguistics (Volume 1: Long Papers)}, pages 631--640,
  Sofia, Bulgaria. Association for Computational Linguistics.

\bibitem[{Li and Ye(2018)}]{li_learning_2018}
Yan Li and Jieping Ye. 2018.
\newblock \href {https://doi.org/10.1145/3219819.3219956} {Learning adversarial
  networks for semi-supervised text classification via policy gradient}.
\newblock In \emph{Proceedings of the 24th {ACM} {SIGKDD} International
  Conference on Knowledge Discovery {\&} Data Mining}, pages 1715--1723. {ACM}.

\bibitem[{Lin et~al.(2017)Lin, Li, He, Sun, and Zhang}]{lin_adversarial_2017}
Kevin Lin, Dianqi Li, Xiaodong He, Ming{-}Ting Sun, and Zhengyou Zhang. 2017.
\newblock \href
  {https://proceedings.neurips.cc/paper/2017/hash/bf201d5407a6509fa536afc4b380577e-Abstract.html}
  {Adversarial ranking for language generation}.
\newblock In \emph{Advances in Neural Information Processing Systems 30: Annual
  Conference on Neural Information Processing Systems 2017}, pages 3155--3165.

\bibitem[{McIntosh and Curran(2009)}]{mcintosh_reducing_2009}
Tara McIntosh and James~R. Curran. 2009.
\newblock \href {https://aclanthology.org/P09-1045} {Reducing semantic drift
  with bagging and distributional similarity}.
\newblock In \emph{Proceedings of the Joint Conference of the 47th Annual
  Meeting of the {ACL} and the 4th International Joint Conference on Natural
  Language Processing of the {AFNLP}}, pages 396--404, Suntec, Singapore.
  Association for Computational Linguistics.

\bibitem[{Paszke et~al.(2019)Paszke, Gross, Massa, Lerer, Bradbury, Chanan,
  Killeen, Lin, Gimelshein, Antiga, Desmaison, K{\"{o}}pf, Yang, DeVito,
  Raison, Tejani, Chilamkurthy, Steiner, Fang, Bai, and
  Chintala}]{paszke_pytorch_2019}
Adam Paszke, Sam Gross, Francisco Massa, Adam Lerer, James Bradbury, Gregory
  Chanan, Trevor Killeen, Zeming Lin, Natalia Gimelshein, Luca Antiga, Alban
  Desmaison, Andreas K{\"{o}}pf, Edward Yang, Zachary DeVito, Martin Raison,
  Alykhan Tejani, Sasank Chilamkurthy, Benoit Steiner, Lu~Fang, Junjie Bai, and
  Soumith Chintala. 2019.
\newblock \href
  {https://proceedings.neurips.cc/paper/2019/hash/bdbca288fee7f92f2bfa9f7012727740-Abstract.html}
  {Pytorch: An imperative style, high-performance deep learning library}.
\newblock In \emph{Advances in Neural Information Processing Systems 32: Annual
  Conference on Neural Information Processing Systems 2019}, pages 8024--8035.

\bibitem[{Qin et~al.(2018)Qin, Xu, and Wang}]{qin_dsgan_2018}
Pengda Qin, Weiran Xu, and William~Yang Wang. 2018.
\newblock \href {https://doi.org/10.18653/v1/P18-1046} {{DSGAN}: Generative
  adversarial training for distant supervision relation extraction}.
\newblock In \emph{Proceedings of the 56th Annual Meeting of the Association
  for Computational Linguistics (Volume 1: Long Papers)}, pages 496--505,
  Melbourne, Australia. Association for Computational Linguistics.

\bibitem[{Ravichandran and Hovy(2002)}]{ravichandran_learning_2002}
Deepak Ravichandran and Eduard Hovy. 2002.
\newblock \href {https://doi.org/10.3115/1073083.1073092} {Learning surface
  text patterns for a question answering system}.
\newblock In \emph{Proceedings of the 40th Annual Meeting of the Association
  for Computational Linguistics}, pages 41--47, Philadelphia, Pennsylvania,
  USA. Association for Computational Linguistics.

\bibitem[{Riloff(1996)}]{riloff_automatically_1996}
Ellen Riloff. 1996.
\newblock Automatically {{Generating Extraction Patterns}} from {{Untagged
  Text}}.
\newblock In \emph{AAAI}, pages 1044--1049.

\bibitem[{Riloff and Jones(1999)}]{riloff_learning_1999}
Ellen Riloff and Rosie Jones. 1999.
\newblock Learning dictionaries for information extraction by multi-level
  bootstrapping.
\newblock In \emph{{{AAAI}}/{{IAAI}}}, pages 474--479.

\bibitem[{Saha et~al.(2017)Saha, Pal, and {Mausam}}]{saha_bootstrapping_2017}
Swarnadeep Saha, Harinder Pal, and {Mausam}. 2017.
\newblock \href {https://doi.org/10.18653/v1/P17-2050} {Bootstrapping for
  numerical open {IE}}.
\newblock In \emph{Proceedings of the 55th Annual Meeting of the Association
  for Computational Linguistics (Volume 2: Short Papers)}, pages 317--323,
  Vancouver, Canada. Association for Computational Linguistics.

\bibitem[{Shi et~al.(2014)Shi, Zhang, Sun, and Han}]{shi_probabilistic_2014}
Bei Shi, Zhenzhong Zhang, Le~Sun, and Xianpei Han. 2014.
\newblock \href {https://aclanthology.org/C14-1215} {A probabilistic
  co-bootstrapping method for entity set expansion}.
\newblock In \emph{Proceedings of {COLING} 2014, the 25th International
  Conference on Computational Linguistics: Technical Papers}, pages 2280--2290,
  Dublin, Ireland. Dublin City University and Association for Computational
  Linguistics.

\bibitem[{Springenberg(2016)}]{springenberg_unsupervised_2016}
Jost~Tobias Springenberg. 2016.
\newblock \href {http://arxiv.org/abs/1511.06390} {Unsupervised and
  semi-supervised learning with categorical generative adversarial networks}.
\newblock In \emph{4th International Conference on Learning Representations}.

\bibitem[{Thelen and Riloff(2002)}]{thelen_bootstrapping_2002}
Michael Thelen and Ellen Riloff. 2002.
\newblock \href {https://doi.org/10.3115/1118693.1118721} {A bootstrapping
  method for learning semantic lexicons using extraction pattern contexts}.
\newblock In \emph{Proceedings of the 2002 Conference on Empirical Methods in
  Natural Language Processing}, pages 214--221. Association for Computational
  Linguistics.

\bibitem[{Tieleman and Hinton(2012)}]{tieleman_lecture_2012}
Tijmen Tieleman and Geoffrey Hinton. 2012.
\newblock Lecture 6.5-rmsprop: Divide the gradient by a running average of its
  recent magnitude.
\newblock \emph{COURSERA: Neural networks for machine learning}, 4(2):26--31.

\bibitem[{Wang et~al.(2019)Wang, Han, Liu, Sun, and Li}]{wang_adversarial_2019}
Xiaozhi Wang, Xu~Han, Zhiyuan Liu, Maosong Sun, and Peng Li. 2019.
\newblock \href {https://doi.org/10.18653/v1/N19-1105} {Adversarial training
  for weakly supervised event detection}.
\newblock In \emph{Proceedings of the 2019 Conference of the North {A}merican
  Chapter of the Association for Computational Linguistics: Human Language
  Technologies, Volume 1 (Long and Short Papers)}, pages 998--1008,
  Minneapolis, Minnesota. Association for Computational Linguistics.

\bibitem[{Williams(1992)}]{williams_simple_1992}
Ronald~J. Williams. 1992.
\newblock Simple statistical gradient-following algorithms for connectionist
  reinforcement learning.
\newblock In \emph{Reinforcement {Learning}}, pages 5--32.

\bibitem[{Yan et~al.(2020{\natexlab{a}})Yan, Han, He, and Sun}]{yan_end_2020}
Lingyong Yan, Xianpei Han, Ben He, and Le~Sun. 2020{\natexlab{a}}.
\newblock \href {https://doi.org/10.1609/aaai.v34i05.6482} {End-to-{End}
  {Bootstrapping} {Neural} {Network} for {Entity} {Set} {Expansion}}.
\newblock In \emph{Proceedings of the {AAAI} {Conference} on {Artificial}
  {Intelligence}}, volume~34, pages 9402--9409.

\bibitem[{Yan et~al.(2020{\natexlab{b}})Yan, Han, He, and
  Sun}]{yan_global_2020}
Lingyong Yan, Xianpei Han, Ben He, and Le~Sun. 2020{\natexlab{b}}.
\newblock \href {https://doi.org/10.18653/v1/2020.findings-emnlp.331} {Global
  bootstrapping neural network for entity set expansion}.
\newblock In \emph{Findings of the Association for Computational Linguistics:
  EMNLP 2020}, pages 3705--3714, Online. Association for Computational
  Linguistics.

\bibitem[{Yan et~al.(2019)Yan, Han, Sun, and He}]{yan_learning_2019}
Lingyong Yan, Xianpei Han, Le~Sun, and Ben He. 2019.
\newblock \href {https://doi.org/10.18653/v1/D19-1028} {Learning to bootstrap
  for entity set expansion}.
\newblock In \emph{Proceedings of the 2019 Conference on Empirical Methods in
  Natural Language Processing and the 9th International Joint Conference on
  Natural Language Processing (EMNLP-IJCNLP)}, pages 292--301, Hong Kong,
  China. Association for Computational Linguistics.

\bibitem[{Yang et~al.(2018)Yang, Chen, Wang, and Xu}]{yang_improving_2018}
Zhen Yang, Wei Chen, Feng Wang, and Bo~Xu. 2018.
\newblock \href {https://doi.org/10.18653/v1/N18-1122} {Improving neural
  machine translation with conditional sequence generative adversarial nets}.
\newblock In \emph{Proceedings of the 2018 Conference of the North {A}merican
  Chapter of the Association for Computational Linguistics: Human Language
  Technologies, Volume 1 (Long Papers)}, pages 1346--1355, New Orleans,
  Louisiana. Association for Computational Linguistics.

\bibitem[{Yangarber et~al.(2002)Yangarber, Lin, and
  Grishman}]{yangarber_unsupervised_2002}
Roman Yangarber, Winston Lin, and Ralph Grishman. 2002.
\newblock \href {https://aclanthology.org/C02-1154} {Unsupervised learning of
  generalized names}.
\newblock In \emph{{COLING} 2002: The 19th International Conference on
  Computational Linguistics}.

\bibitem[{Yoshida et~al.(2010)Yoshida, Ikeda, Ono, Sato, and
  Nakagawa}]{yoshida_person_2010}
Minoru Yoshida, Masaki Ikeda, Shingo Ono, Issei Sato, and Hiroshi Nakagawa.
  2010.
\newblock \href {https://doi.org/10.1145/1835449.1835454} {Person name
  disambiguation by bootstrapping}.
\newblock In \emph{Proceeding of the 33rd International {ACM} {SIGIR}
  Conference on Research and Development in Information Retrieval}, pages
  10--17. {ACM}.

\bibitem[{Yu et~al.(2017)Yu, Zhang, Wang, and Yu}]{yu_seqgan_2017}
Lantao Yu, Weinan Zhang, Jun Wang, and Yong Yu. 2017.
\newblock \href {http://aaai.org/ocs/index.php/AAAI/AAAI17/paper/view/14344}
  {Seqgan: Sequence generative adversarial nets with policy gradient}.
\newblock In \emph{Proceedings of the Thirty-First {AAAI} Conference on
  Artificial Intelligence}, pages 2852--2858. {AAAI} Press.

\bibitem[{Zhang et~al.(2020)Zhang, Shen, Shang, and Han}]{zhang_empower_2020}
Yunyi Zhang, Jiaming Shen, Jingbo Shang, and Jiawei Han. 2020.
\newblock \href {https://doi.org/10.18653/v1/2020.acl-main.725} {Empower entity
  set expansion via language model probing}.
\newblock In \emph{Proceedings of the 58th Annual Meeting of the Association
  for Computational Linguistics}, pages 8151--8160, Online. Association for
  Computational Linguistics.

\bibitem[{Zupon et~al.(2019)Zupon, Alexeeva, Valenzuela-Esc{\'a}rcega, Nagesh,
  and Surdeanu}]{zupon_lightly-supervised_2019}
Andrew Zupon, Maria Alexeeva, Marco Valenzuela-Esc{\'a}rcega, Ajay Nagesh, and
  Mihai Surdeanu. 2019.
\newblock \href {https://doi.org/10.18653/v1/W19-1504} {Lightly-supervised
  representation learning with global interpretability}.
\newblock In \emph{Proceedings of the Third Workshop on Structured Prediction
  for {NLP}}, pages 18--28, Minneapolis, Minnesota. Association for
  Computational Linguistics.

\end{thebibliography}
}

\end{document}